\documentclass{article}

\usepackage{amsmath}
\usepackage{wrapfig}
\usepackage{float}
\usepackage{amssymb}
\usepackage{graphicx}
\usepackage{booktabs}
\usepackage[table]{xcolor}
\usepackage{colortbl}

\definecolor{lightred}{HTML}{FFAAAA}
\usepackage[preprint]{corl_2026} 

\title{MimicIK: Real-Time Generative Inverse Kinematics from Teleoperation with FK Consistency}
\author{
{\small
\begin{tabular}{c}
Jiahao Yang$^{1}$,
Shenhao Yan$^{1}$,
Fan Feng$^{1}$,
Chengsi Yao$^{1}$,
Ge Wang$^{1,2}$,\\
Zhixin Mai,
Yiming Zhao$^{1}$,
and Yatong Han$^{\dagger,1}$\\[0.5em]
$^{1}$Ising AI\\
$^{2}$CUHK-Shenzhen\\
$^{\dagger}$corresponding author: \texttt{rstanten@alumni.stanford.edu}
\end{tabular}
}
}

\begin{document}
\maketitle


\begin{abstract}
     Inverse kinematics (IK) remains a critical bottleneck for real-time robot manipulation. Classical numerical solvers achieve high geometric precision but often suffer from discontinuous branch switching and unstable behavior near kinematic singularities during closed-loop deployment. Meanwhile, learned IK approaches frequently struggle to balance spatial accuracy, motion smoothness, and real-time efficiency, particularly when trained on noisy human teleoperation data. We present \textbf{MimicIK}, a real-time generative inverse kinematics framework that learns smooth and robust joint-space motion priors from teleoperation demonstrations through conditional flow matching. Given the current joint configuration and a target end-effector pose, MimicIK predicts continuous delta-joint commands using an efficient two-step iterative refinement process based on a Minimal Iterative Policy (MIP) backbone. To enforce physical consistency, we further introduce an FK consistency loss, a differentiable forward-kinematics regularization that penalizes task-space deviations from the target pose during training. We evaluate MimicIK on a real-world 6-DOF robot dataset containing 8,848 teleoperation demonstrations. MimicIK achieves a mean position error of 4.65 mm, a 10 mm success rate of 92.01\%, and a trajectory spike rate of only 7.99\%. Compared with a UNet diffusion baseline, our method improves both spatial accuracy and motion smoothness while reducing inference latency from 21.66 ms to 6.74 ms. Furthermore, unlike deterministic MLP baselines that catastrophically diverge under out-of-distribution deployment, MimicIK remains stable near singular configurations and enables robust 20 Hz real-time control on deployment hardware.
\end{abstract}

\vspace{-0.6em}
\begin{figure}[H]
    \centering
    \includegraphics[width=0.95\textwidth]{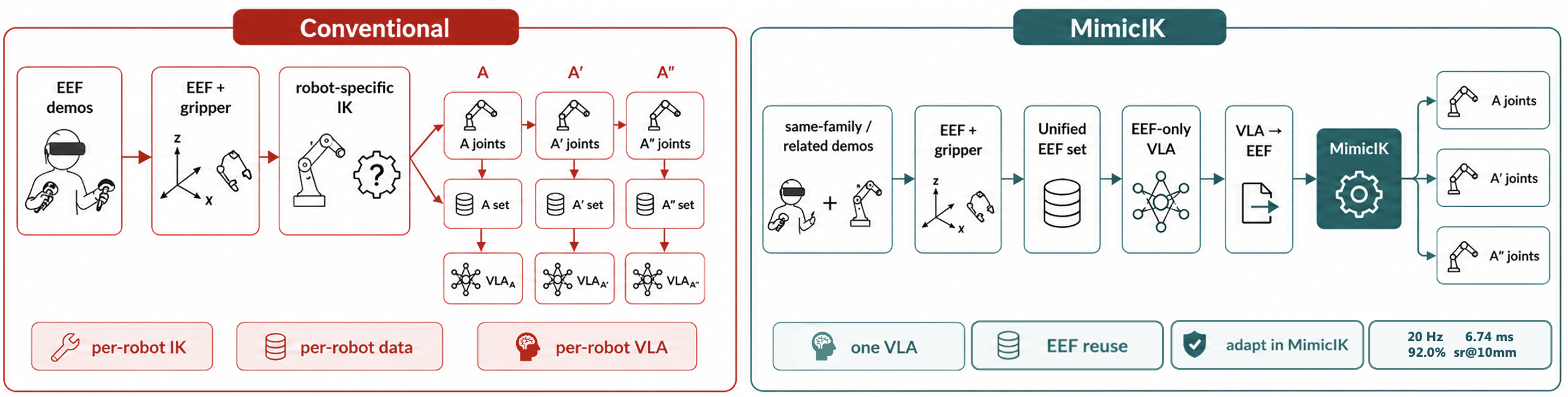}
    \vspace{-0.8em}
    \caption{\textbf{Paradigm Shift in EEF-Level Robot Data Scaling.}
    \textbf{Left:} Conventional robot-specific IK conversion fragments shared EEF demonstrations into isolated joint-space datasets.
    \textbf{Right:} MimicIK keeps VLA training in embodiment-agnostic EEF space and performs robot-specific real-time IK adaptation only at deployment.}
    \label{fig:teaser}
    \vspace{-1.0em}
\end{figure}

 

\section{Introduction}
Inspired by the massive flywheel effects demonstrated by large language models on internet-scale data, the embodied intelligence community is increasingly relying on scaling laws to advance Vision-Language-Action (VLA) models~\cite{brohan2022rt,zitkovich2023rt}. However, acquiring real-world robotic data remains prohibitively expensive, as traditional teleoperation requires costly hardware and suffers from low collection efficiency. To overcome this data bottleneck, paradigm shifts such as the Universal Manipulation Interface (UMI)~\cite{chi2024universal} have emerged. By utilizing handheld or wearable devices to collect spatial trajectories in $SE(3)$, visual observations, and gripper states, researchers can bypass the constraints of specific robot hardware and construct large-scale cross-embodiment datasets.

While embodiment-agnostic data collection paves the way for scaling robot learning, deploying end-effector (EEF) trajectories onto heterogeneous robot arms introduces a critical challenge. A scalable approach is to train VLA models to output EEF actions and rely on real-time Inverse Kinematics (IK) solvers during deployment to map these $SE(3)$ targets into robot-specific joint configurations. Consequently, real-time IK becomes a key computational bottleneck in the EEF-level VLA pipeline. Classical numerical solvers, such as Jacobian-based methods~\cite{buss2004introduction}, often struggle near kinematic singularities, resulting in erratic movements, deadlocks, or explosive joint rotations during closed-loop control~\cite{chiaverini2002singularity}. In contrast, human operators can intuitively and smoothly navigate robot arms out of such kinematic dead-ends. This contrast motivates a data-driven generative paradigm: by modeling human demonstrations of singularity recovery, we aim to inject natural ``self-rescuing'' priors into the IK solver.

To bypass numerical instabilities and leverage the rich motion priors embedded in UMI-style demonstrations, data-driven generative approaches have gained traction~\cite{zhao2023learning, mandlekar2021matters}. However, human teleoperation data is inherently noisy, multimodal, and non-smooth. Standard regression models, such as MLPs, suffer from mode collapse and fail to capture the multimodality of human actions~\cite{florence2022implicit}. Although recent generative paradigms such as Diffusion Models excel at modeling complex action distributions~\cite{chi2025diffusion, pearce2023imitating}, they introduce two critical limitations for real-time IK: prohibitive inference latency due to dozens of iterative denoising steps~\cite{song2020denoising}, and a tendency to overfit high-frequency hand tremors in teleoperation data, leading to jerky joint trajectories.

To overcome the dichotomy between physical accuracy and inference efficiency in the VLA era, we propose MimicIK, a generative IK framework that distills human-like motion priors into a robust real-time solver. We formulate continuous joint displacement prediction as a conditional flow-matching problem~\cite{lipman2022flow, liu2022flow}. By leveraging the simple transport structure of flow matching, MimicIK learns smooth joint-space transitions and enables a highly efficient two-step inference process. Furthermore, recognizing that generative models may lack strict adherence to physical constraints, we introduce an FK consistency loss, a differentiable forward-kinematics penalty applied during training. This loss structurally encourages the generated joint configuration to match the target end-effector pose, improving spatial accuracy while preserving smooth motion priors.

In summary, we present MimicIK, a mimicry-based inverse kinematics framework built upon conditional flow matching, which learns smooth and human-like joint priors from noisy teleoperation demonstrations while requiring only a highly efficient two-step inference process. To align the learned generative motion prior with physical kinematic constraints, we introduce a differentiable forward-kinematics consistency loss, which substantially improves spatial accuracy and prevents training divergence without compromising trajectory smoothness. Extensive experiments on a real-world 6-DOF robotic dataset containing 8,848 teleoperation demonstrations show that MimicIK achieves a mean position error of 4.65 mm and limits the spike rate to 7.99\%, significantly outperforming diffusion-based and MLP baselines. With an inference latency of only 6.74 ms and an energy cost of 398 mJ per step, MimicIK supports robust 20 Hz real-time control and successfully escapes singular traps on physical hardware without causing explosive joint rotations.


\section{Related Work}
\label{sec:related work}

\subsection{Scalable Robot Learning and Embodiment-Agnostic Data}
Recent progress in manipulation has been heavily driven by large-scale datasets~\citep{dasari2019robonet,walke2023bridgedata,khazatsky2024droid} and high-capacity Vision-Language-Action (VLA) models such as RT-1/2, Open X-Embodiment, Octo, OpenVLA, and $\pi_0$~\citep{brohan2022rt,zitkovich2023rt,o2024open,mees2024octo,kim2024openvla,black2024pi0}. However, collecting robot-specific demonstrations remains prohibitively expensive. To mitigate this, embodiment-agnostic frameworks like the Universal Manipulation Interface (UMI)~\citep{chi2024universal} and Real2Render2Real~\citep{yu2025real2render2real} collect task data directly in the end-effector (EEF) space. While this greatly facilitates cross-embodiment scaling, deploying these EEF policies on heterogeneous real-world arms requires online conversion into joint commands. MimicIK addresses this critical deployment bottleneck, acting as a real-time adapter that bridges scalable, embodiment-agnostic VLA outputs with embodiment-specific joint control.

\subsection{From Classical to Generative Inverse Kinematics}
Classical numerical IK solvers—including Jacobian-based methods, TRAC-IK~\citep{beeson2015trac}, Pinocchio~\citep{carpentier2019pinocchio}, cuRobo~\citep{sundaralingam2023curobo}, and PyBullet~\citep{coumans2016pybullet}—are highly precise and widely adopted. However, they struggle with the one-to-many mapping of redundant manipulators. Near kinematic singularities, they often produce discontinuous joint jumps or severe branch switching across consecutive timesteps, as they fundamentally lack human-like motion priors. To capture this multimodality, learning-based approaches like IKFlow~\citep{ames2022ikflow} and recent diffusion IK models~\citep{zhang2025ikdiffuser} sample diverse valid joint configurations. Unlike these methods that primarily solve isolated spatial queries, MimicIK formulates IK as a local, history-aware delta-joint prediction problem. This sequential formulation ensures the temporal continuity essential for executing continuous VLA trajectories on real hardware.

\subsection{Efficient Generative Policies and Kinematic Regularization}
Diffusion Policy~\citep{chi2025diffusion} and Flow Matching algorithms~\citep{braun2024riemannian,lipman2022flow} have proven highly effective at modeling complex, multimodal action distributions. Yet, their iterative denoising mechanisms incur prohibitive latency, making them unsuited for high-frequency (e.g., 20Hz+) real-time IK control loops. Recent findings in \textit{Much Ado About Noising}~\citep{pan2025much} demonstrate that a Minimal Iterative Policy (MIP) can recover generative control performance via lightweight, two-step supervised iterative computation. MimicIK explicitly adopts this MIP backbone to achieve rapid generative inference. Furthermore, to prevent the generative model from hallucinating physically unviable configurations, we integrate a differentiable forward kinematics (FK)~\citep{molschl2023differentiable} consistency loss during training. This structural regularization securely anchors the learned human motion priors to the strict rigid-body geometry of the robot, ensuring safety without sacrificing inference speed.


\section{MimicIK}
\label{sec:methodology}

	MimicIK is designed as a real-time generative inverse kinematics decoder that maps end-effector commands to smooth joint-space actions. Unlike conventional IK solvers that treat each target pose as an independent geometric optimization problem, MimicIK learns a data-driven joint-space motion prior from real robot trajectories while remaining constrained by differentiable forward kinematics. Our final deployed model is MIP+FK(0.1), which combines Minimal Iterative Policy-based flow matching with an auxiliary FK consistency loss weighted by 0.1.
    \begin{figure*}[t]
    \centering
    \includegraphics[width=\textwidth]{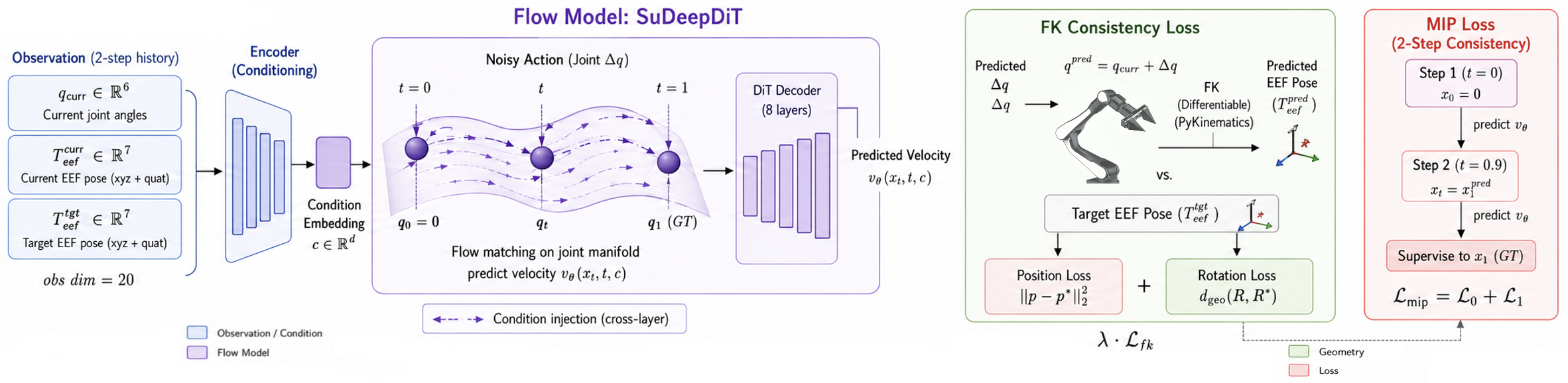}
    \caption{\textbf{Architecture of the MimicIK Framework.} The model receives an observation history comprising the current joint configuration ($q_{\mathrm{curr}}$), current end-effector pose ($x_{\mathrm{curr}}$), and target end-effector pose ($x_{\mathrm{tgt}}$). This condition is injected into a pose-conditioned delta-joint flow generator (parameterized by a SuDeepDiT Transformer), which decodes the continuous joint displacement ($\Delta q$) via a highly efficient two-step inference process. During training, the framework is optimized using two primary IK-aware objectives: a joint-space teleoperation prior ($\mathcal{L}_{\mathrm{MIP}}$) distilled from human demonstrations, and a task-space differentiable forward kinematics (FK) consistency loss ($\mathcal{L}_{\mathrm{FK}}$) that structurally penalizes spatial deviations from the target rigid-body geometry.}
    \label{fig:method_architecture}
\end{figure*}

\subsection{Problem Formulation and Data Representation}
We consider the inverse kinematics problem in a closed-loop robot deployment setting. At each control step, the robot observes its current joint configuration $q_{\mathrm{curr}} \in \mathbb{R}^{d_q}$, and receives a target end-effector pose $x_{\mathrm{tgt}} \in SE(3)$, where $x_{\mathrm{tgt}}$ is represented by a Cartesian position and a unit quaternion. The goal is to predict a joint-space displacement
$$
\Delta q = q_{\mathrm{tgt}} - q_{\mathrm{curr}}
$$
such that the resulting joint command $q_{\mathrm{cmd}} = q_{\mathrm{curr}} + \Delta q$ reaches the desired end-effector pose under the robot's forward kinematics:
$$
FK(q_{\mathrm{cmd}}) \approx x_{\mathrm{tgt}}
$$
Rather than directly predicting absolute joint angles, we predict $\Delta q$. This local formulation is more suitable for real-time deployment because it conditions the prediction on the current robot state and naturally encourages temporal continuity. It also reduces the ambiguity of the IK mapping: multiple absolute joint configurations may correspond to similar end-effector poses, but the desired local displacement is better constrained by the current configuration.

For each frame, the model input is constructed as $o = [q_{\mathrm{curr}}, x_{\mathrm{curr}}, x_{\mathrm{tgt}}]$, where $x_{\mathrm{curr}}$ is the current end-effector pose. In our implementation, $q_{\mathrm{curr}}$ is a 6-DoF joint vector and both $x_{\mathrm{curr}}$ and $x_{\mathrm{tgt}}$ are represented as 7-dimensional pose vectors consisting of position and quaternion. Therefore, each frame-level observation is 20-dimensional, $o \in \mathbb{R}^{20}$. The supervised target is the delta-joint command $a = \Delta q \in \mathbb{R}^{6}$.

We train on real robot trajectories collected from previous manipulation tasks. These trajectories contain the motion preferences induced by human demonstrations, including how the arm moves through redundant regions and how it avoids abrupt posture changes. However, real demonstration data is not perfectly kinematically clean. At the microscopic frame level, the data may contain high-frequency noise, sensor delay, and small action-pose mismatches. This motivates a method that does not simply memorize joint labels, but also regularizes the learned motion distribution using strict robot kinematics.

\subsection{Generative IK via Flow Matching}

The IK mapping is inherently multi-modal. Deterministic regression models tend to average over these modes, producing valid-looking but suboptimal joint commands. Classical numerical IK solvers avoid learning this distribution but are sensitive to initialization and can abruptly jump between different solution branches near singular configurations. We therefore formulate IK as conditional generative modeling.

Given the observation $o$, our model learns a conditional distribution over delta-joint actions,
$p_{\theta}(\Delta q \mid q_{\mathrm{curr}}, x_{\mathrm{curr}}, x_{\mathrm{tgt}})$,
using a flow-matching formulation. Let $x_1$ denote the ground-truth delta-joint action sampled from the dataset, and let $x_0$ denote a noise sample drawn from a simple prior distribution. A continuous interpolation between the two is defined as:
$$
x_s = (1-s)x_0 + sx_1, \quad s \in [0,1]
$$
The model learns a conditional velocity field $v_{\theta}(x_s, o, s)$ that transports noisy samples toward the demonstrated action distribution. The corresponding flow-matching objective is:
$$
\mathcal{L}_{\mathrm{FM}} =
\mathbb{E}_{s,x_0,x_1,o}
\left[
\left\|
v_{\theta}(x_s,o,s) - (x_1-x_0)
\right\|_2^2
\right]
$$

In principle, the learned flow can be solved through many iterative denoising or ODE integration steps. However, high-frequency robot control requires an IK decoder with extremely low inference latency. Standard DDPM-style diffusion policies typically require dozens of iterative updates per query, making them poorly suited for 20 Hz real-time IK deployment.

To enable efficient deployment, we adopt the Minimal Iterative Policy (MIP)~\cite{pan2025much} as a lightweight iterative realization of the flow-based generative policy. Rather than numerically integrating the continuous flow trajectory with many refinement steps, MIP directly performs a small number of supervised iterative updates in the action space. Given an initial noisy action estimate $x_0$, the policy predicts a refined delta-joint action through:
$$
\hat{x}_{1}
=
x_0 + \alpha v_{\theta}(x_0,o,s)
$$
where $\alpha$ denotes the integration step size. In our implementation, only two refinement iterations are used to produce the final delta-joint prediction:
$$
\Delta \hat{q} = \hat{x}_{1}
$$

Following MIP, we supervise the model at two refinement stages rather than relying on long denoising chains. We denote the resulting two-step supervised flow objective as $\mathcal{L}_{\mathrm{MIP}}$.This lightweight iterative formulation preserves the expressive generative modeling capability of flow-based policies while drastically reducing inference latency. Consequently, MimicIK can capture non-trivial joint-space motion priors from teleoperation demonstrations without incurring the computational overhead of long diffusion-style sampling. During deployment, the upstream VLA policy only needs to output target end-effector poses, while MimicIK performs fast real-time end-effector-to-joint decoding.(Please refer to Appendix A for detailed network architectures and training hyperparameters).

\subsection{Differentiable FK Consistency Loss}
Although flow matching learns the demonstrated joint-space motion distribution, pure imitation of $\Delta q$ is not sufficient for reliable IK deployment due to inherent sensor noise and calibration errors in real robot data. To constrain the learned generator with robot geometry, we introduce a differentiable forward-kinematics consistency loss. 

Given the predicted delta-joint command $\Delta \hat{q}$, we first compute the predicted joint command $\hat{q}_{\mathrm{cmd}} = q_{\mathrm{curr}} + \Delta \hat{q}$, and pass it through the robot's differentiable forward-kinematics model:
$$
\hat{x} = FK(\hat{q}_{\mathrm{cmd}})
$$
The predicted pose $\hat{x}$ is compared against the target pose $x_{\mathrm{tgt}}$. For the translational component, we use an $L_2$ position loss:
$$
\mathcal{L}_{\mathrm{pos}} = \left\| \hat{p} - p_{\mathrm{tgt}} \right\|_2^2
$$
For the rotational component, we measure the discrepancy using a rotation distance:
$$
\mathcal{L}_{\mathrm{rot}} = d_R(\hat{R}, R_{\mathrm{tgt}})
$$
The full FK consistency loss is $\mathcal{L}_{\mathrm{FK}} = \mathcal{L}_{\mathrm{pos}} + \lambda_{\mathrm{rot}}\mathcal{L}_{\mathrm{rot}}$. The final training objective of our deployed model is:
\[ \mathcal{L}_{\mathrm{total}} = \mathcal{L}_{\mathrm{MIP}} + \lambda_{\mathrm{FK}}\mathcal{L}_{\mathrm{FK}}, \quad \lambda_{\mathrm{FK}} = 0.1. \]

The FK consistency term plays a crucial role: it pulls the generative model back into the rigid-body kinematic structure of the robot. Importantly, the FK loss is used during training only. At test time, MimicIK does not perform iterative FK optimization. The deployed solver remains a fast feed-forward generative IK decoder. Empirically, through a systematic ablation study across $\lambda \in [0.05, 0.5]$, we find that an FK weight of $\lambda=0.1$ provides the optimal trade-off between geometric accuracy and joint-space smoothness (see \textbf{Appendix B} for detailed quantitative comparisons).


\section{Experiment}
\label{sec:experiment}

	Our experiments are designed to answer three key questions: \textbf{(1)} Can MimicIK achieve the spatial accuracy of numerical solvers while maintaining the kinematic smoothness necessary for real-world tasks? \textbf{(2)} How does the proposed FK consistency loss impact training stability? \textbf{(3)} Does the framework satisfy the stringent latency and robustness requirements for high-frequency hardware deployment under out-of-distribution (OOD) scenarios?

\subsection{Experimental Setup and Dataset}
To train and evaluate MimicIK, we utilized a large-scale, real-robot manipulation dataset collected via human teleoperation on an AIRBOT dual-arm platform. The data was recorded at 20 Hz in a LeRobot Parquet format\citep{cadene2024lerobot}, with the end-effector Cartesian workspace spanning approximately 0.43 m $\times$ 0.40 m $\times$ 0.23 m. 

To rigorously test real-world robustness and prevent data leakage, we explicitly held out a portion of the teleoperation tasks to form the test set, separating the data by distinct recording sessions rather than random shuffling. This design ensures the evaluation reflects true cross-session generalization under realistic distribution shifts. Specifically, the Training Set contains 6,427 trajectories (approx. 23.9 hours of motion) with an average duration of 13.4 seconds per episode. Crucially, this corpus explicitly incorporates dedicated singularity recovery demonstrations (approx. 400 episodes), where human operators manually guided the robot out of induced kinematic traps. The Held-out Test Set comprises 2,418 entirely unseen trajectories (approx. 9.2 hours). Furthermore, to evaluate deployment-critical failure modes such as Cartesian spikes and solver divergence, we collected a highly challenging OOD Robustness Set consisting of 3 long-horizon trajectories (approx. 21 minutes, averaging 415 s/episode) that intentionally traverse near-singular configurations.

\begin{figure}[t]
    \centering
    \begin{minipage}{0.31\linewidth}
        \centering
        \includegraphics[width=\linewidth]{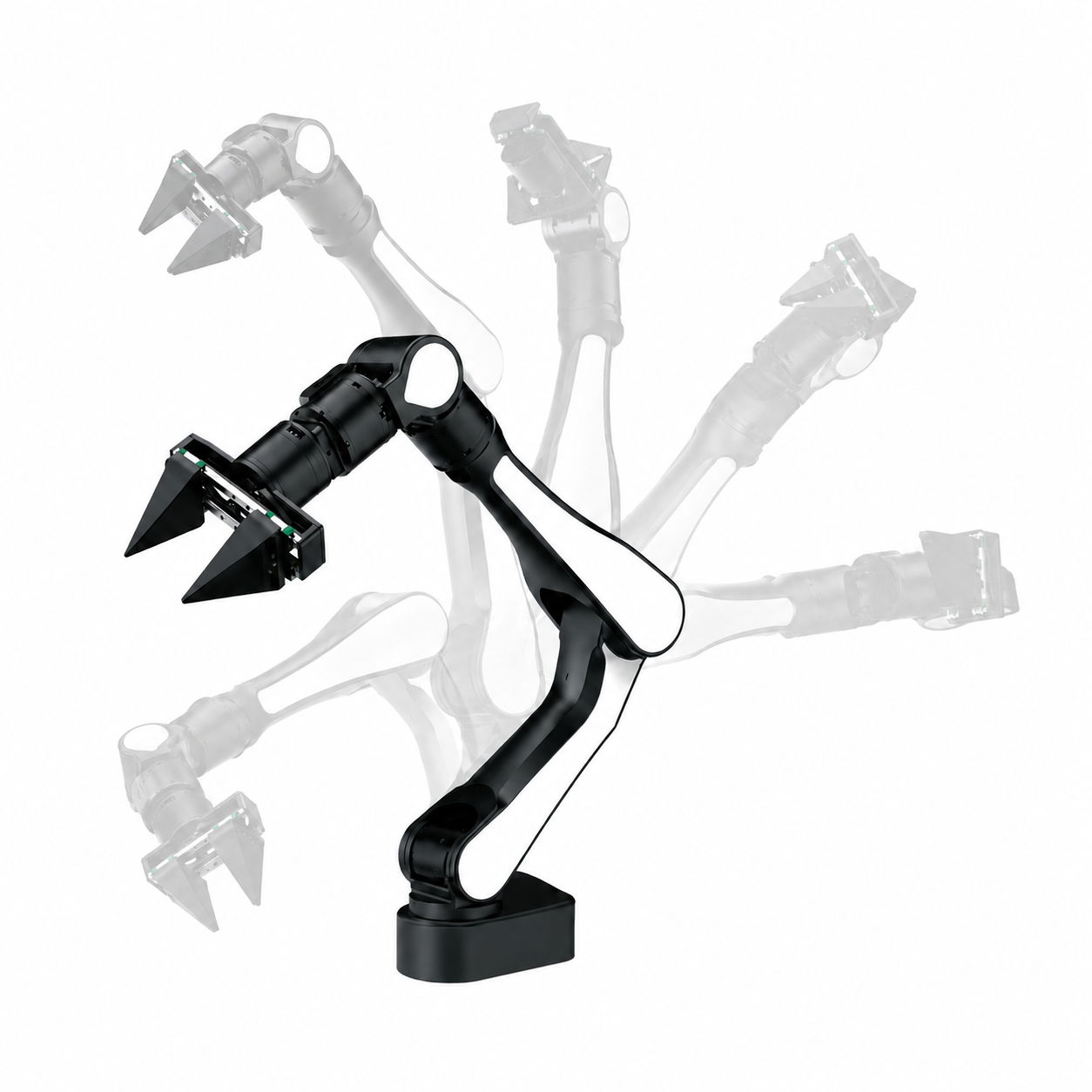}
        \centerline{\small (a) AIRBOT Platform}
    \end{minipage}
    \hspace{0.025\linewidth}
    \begin{minipage}{0.52\linewidth}
        \centering
        \includegraphics[width=\linewidth]{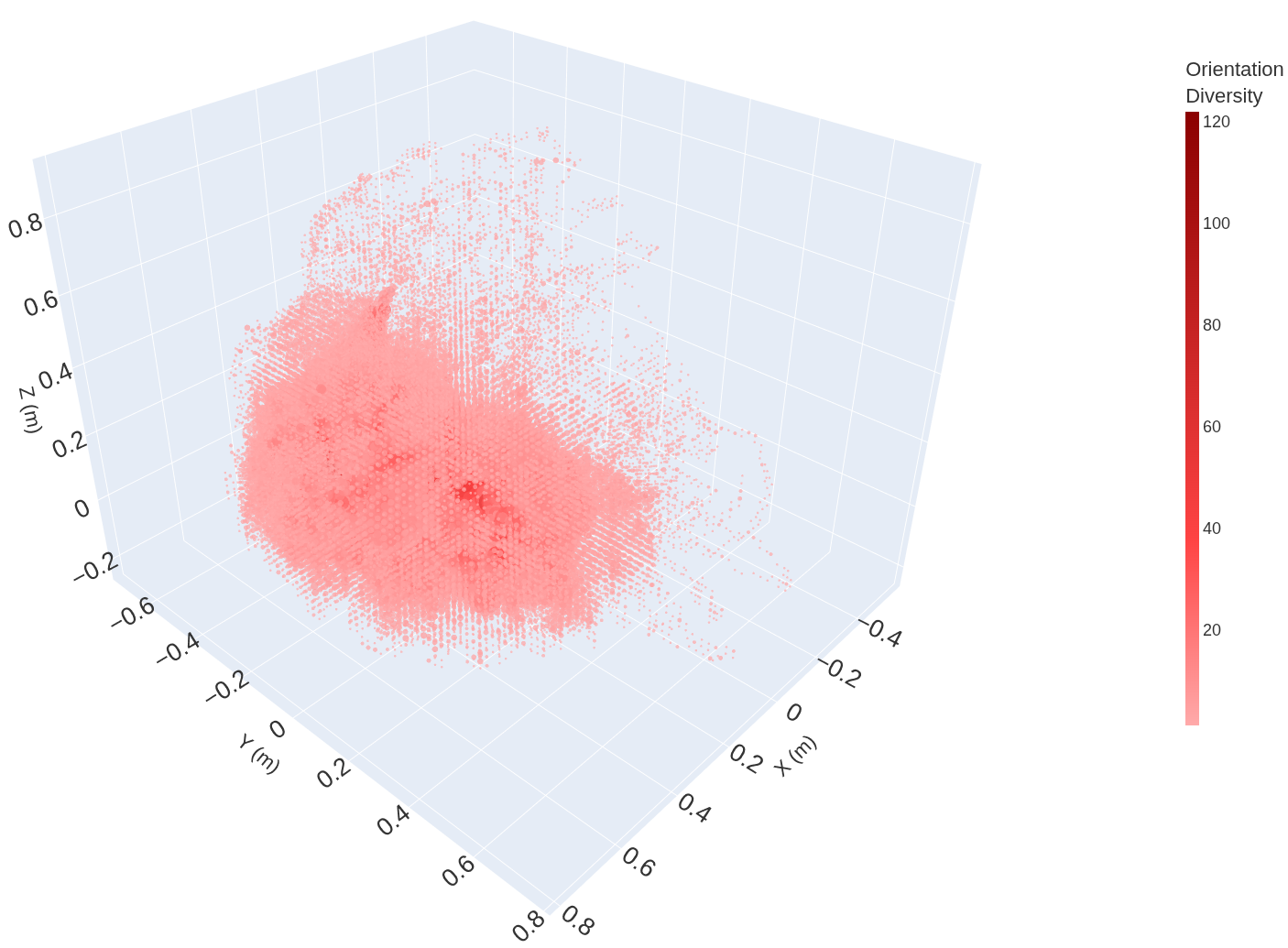}
        \centerline{\small (b) EEF Workspace Coverage}
    \end{minipage}

    \vspace{0.05cm}
    \caption{\textbf{Hardware Setup and Dataset Distribution.} (a) The AIRBOT dual-arm platform used for human teleoperation data collection. (b) The 3D heatmap visualizes the extensive spatial distribution of the 8,848 real-robot trajectories. Color intensity indicates orientation diversity within each voxel, highlighting the rich, multi-modal human motion priors captured in our dataset.}
    \label{fig:hardware_and_dataset}
\end{figure}
\subsection{Baselines and Metrics}
We compare MimicIK (deployed as a 2-step Minimal Iterative Policy solver with $\lambda_{\mathrm{FK}}=0.1$) against three classes of baselines via a unified IK Expert Server:

\noindent\textbf{Generative Baseline (UNet Diffusion):} A standard 10-step DDPM-based\citep{ho2020denoising} diffusion policy.\\
\textbf{Regression Baseline (Pure MLP):} A deterministic 1-step feed-forward network.\\
\textbf{Traditional Solvers:} Pinocchio (Jacobian-based), PyBullet (Damped Least Squares), and cuRobo (state-of-the-art GPU optimization).
\begin{table*}[t]
\centering
\caption{\textbf{Main quantitative results on the held-out test set and OOD deployment set.}
MimicIK achieves the best trade-off between spatial accuracy, trajectory smoothness, inference latency, and deployment robustness. Results for learned methods are averaged over 3 random seeds.}
\label{tab:main_results}

\renewcommand{\arraystretch}{1.08}

\resizebox{\textwidth}{!}{%
\begin{tabular}{lccccccc}
\toprule
\textbf{Method} &
\textbf{Mean Err.} &
\textbf{P95} &
\textbf{SR@10mm} &
\textbf{Spike} &
\textbf{Latency} &
\textbf{Power} &
\textbf{OOD} \\
&
\textbf{(mm) $\downarrow$} &
\textbf{(mm) $\downarrow$} &
\textbf{(\%) $\uparrow$} &
\textbf{(\%) $\downarrow$} &
\textbf{(ms) $\downarrow$} &
\textbf{(W) $\downarrow$} &
\\
\midrule

Pure MLP
& $4.53 \pm 0.62$
& $17.47 \pm 0.94$
& 89.09
& 10.91
& \textbf{0.11}
& \textbf{55.6}
& \textcolor{red}{Diverged} \\

UNet Diffusion
& $5.89 \pm 0.63$
& $19.48 \pm 2.66$
& 85.59
& 14.41
& 21.66
& 158.2
& Bounded \\

\midrule

MimicIK (no-FK)
& $6.29 \pm 2.63$
& $14.01 \pm 2.10$
& 78.11
& 21.89
& 6.79
& 59.2
& Bounded \\

\textbf{MimicIK (Ours)}
& $\mathbf{4.65 \pm 0.35}$
& $\mathbf{11.16 \pm 1.08}$
& \textbf{92.01}
& \textbf{7.99}
& 6.74
& 58.9
& \textbf{Bounded} \\

\midrule

Pinocchio
& \textbf{0.005}
& \textbf{0.024}
& \textbf{100.0}
& --
& 0.03
& --
& Singular \\

cuRobo
& $\approx 0.00$
& $\approx 0.00$
& 99.76
& --
& 1.55
& --
& Singular \\

PyBullet
& 1.69
& 5.63
& 98.20
& --
& 0.91
& --
& Singular \\

\bottomrule
\end{tabular}%
}

\vspace{0.35em}

\begin{minipage}{0.98\textwidth}
\footnotesize
\textit{Note:}
Traditional IK solvers achieve extremely low offline pose error, but frequently exhibit discontinuous branch switching or singular traps during closed-loop OOD deployment.
Spike rate is reported only for learned policies.
\end{minipage}
\vspace{-5mm}
\end{table*}
\vspace{-5mm}
\subsection{Quantitative Results and Computational Efficiency}
\vspace{-2mm}
Table~\ref{tab:main_results} summarizes the comprehensive performance across all metrics. MimicIK achieves a mean position error of 4.65 mm and an SR@10mm of 92.01\%, significantly outperforming the UNet Diffusion baseline (5.89 mm). 

\textbf{Real-Time Inference and GPU Utilization:} A critical bottleneck for deploying Vision-Language-Action (VLA) models is the tight latency budget of high-frequency control. For a standard 20 Hz robotic system, the IK solver must consume only a fraction of the 50 ms budget to leave sufficient time for upstream vision encoders. The standard UNet Diffusion baseline requires 21.66 ms per inference and demands a massive GPU power load of 158.2 W. In contrast, MimicIK drastically reduces this latency to just 6.74 ms (a 3.2$\times$ speedup) and operates at a much lighter GPU power of 58.9 W. When factoring in the inference time, MimicIK requires only 398 mJ per step compared to the massive 3,424 mJ consumed by diffusion—an 8.6$\times$ reduction in energy footprint. This extraordinary efficiency makes MimicIK uniquely suited as a lightweight generative IK backbone for edge deployments.

\begin{wrapfigure}{r}{0.55\linewidth}
    \centering
    \includegraphics[width=\linewidth]{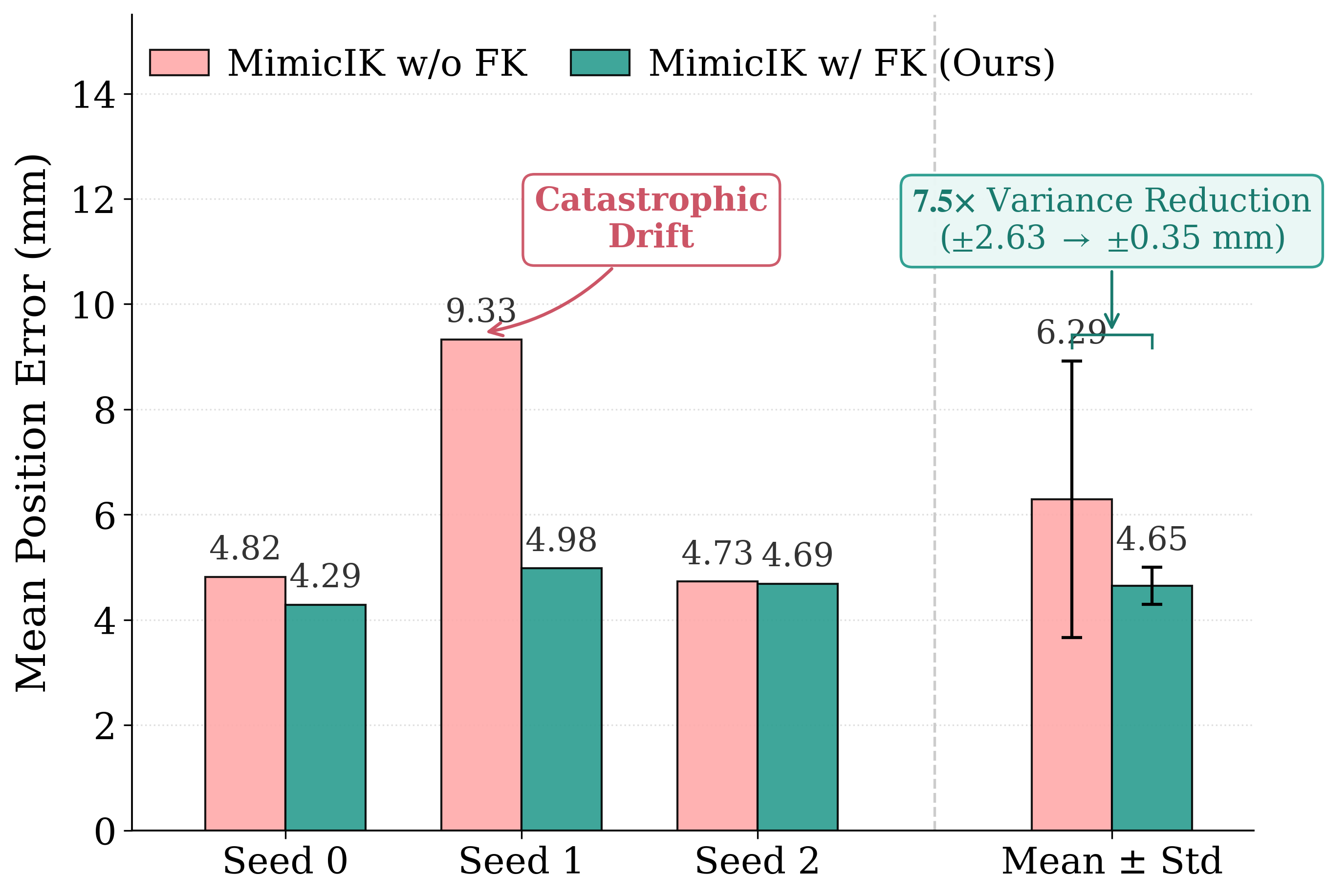}
    \caption{\textbf{FK consistency loss improves training stability.}
    Across three seeds, FK regularization prevents catastrophic drift and reduces cross-seed variance by 7.5$\times$.}
    \label{fig:fk_ablation}
\end{wrapfigure}

\textbf{The Crucial Role of FK Consistency:} Ablation on the FK loss reveals that its primary contribution is acting as a powerful structural regularizer that guarantees \textit{training stability} (Figure~\ref{fig:fk_ablation}). Without FK supervision, pure joint-space flow matching is prone to learning physically unconstrained manifolds. This leads to severe training variance and catastrophic divergence in certain initialization seeds (e.g., the no-FK model in Seed 1 drifts to a high error of 9.33 mm). By introducing the differentiable FK penalty, the generative model is structurally anchored to the physical rigid-body workspace. This regularization effectively rescues divergent seeds, reducing the cross-seed standard deviation by a massive 7.5$\times$ (from $\pm$ 2.63 mm to $\pm$ 0.35 mm) and improving the overall mean error by 26\%. 

Furthermore, beyond ensuring convergence, the FK loss provides a secondary benefit of fine-grained spatial refinement. For optimally converged seeds (e.g., Seed 0), the end-effector task-space gradients complement the joint-space objective, directly pushing the local position error from 4.82 mm down to 4.29 mm. Consequently, the FK-regularized policy strictly adheres to the demonstrated trajectories, ensuring both robust spatial accuracy and physical viability for closed-loop execution.

\subsection{Real-Robot Deployment and OOD Robustness}
While offline metrics provide a baseline, physical deployment under out-of-distribution (OOD) and near-singular configurations is the true test of an IK solver (Table~\ref{tab:main_results}).

\textbf{MLP Fragility:} Despite competitive offline metrics, the Pure MLP is unsafe for physical deployment. Under OOD evaluation, all MLP seeds catastrophically diverged (outputting NaNs) within 300 frames, which would immediately trigger over-velocity limits and severe hardware crashes.

\textbf{Traditional Solver Artifacts:} While numerically precise offline, solvers like cuRobo and PyBullet lack human-like motion priors. During physical execution, they frequently fall into singularity traps, occasionally commanding sudden, erratic 180-degree wrist flips that risk hardware collisions and cable tangling.

\textbf{MimicIK's Smooth Deployment:} Conversely, MimicIK gracefully handles OOD configurations. Distilled from human teleoperation---including the recovery demonstrations from Section 4.1---MimicIK inherently understands how to navigate out of kinematic dead-ends. It seamlessly escapes singular traps with strict temporal smoothness, avoiding the explosive rotations that plague numerical methods (Figure~\ref{fig:real_robot_deployment}; see \textbf{Appendix C} and supplementary video).

\begin{figure}[H]
    \centering
    \includegraphics[width=\columnwidth]{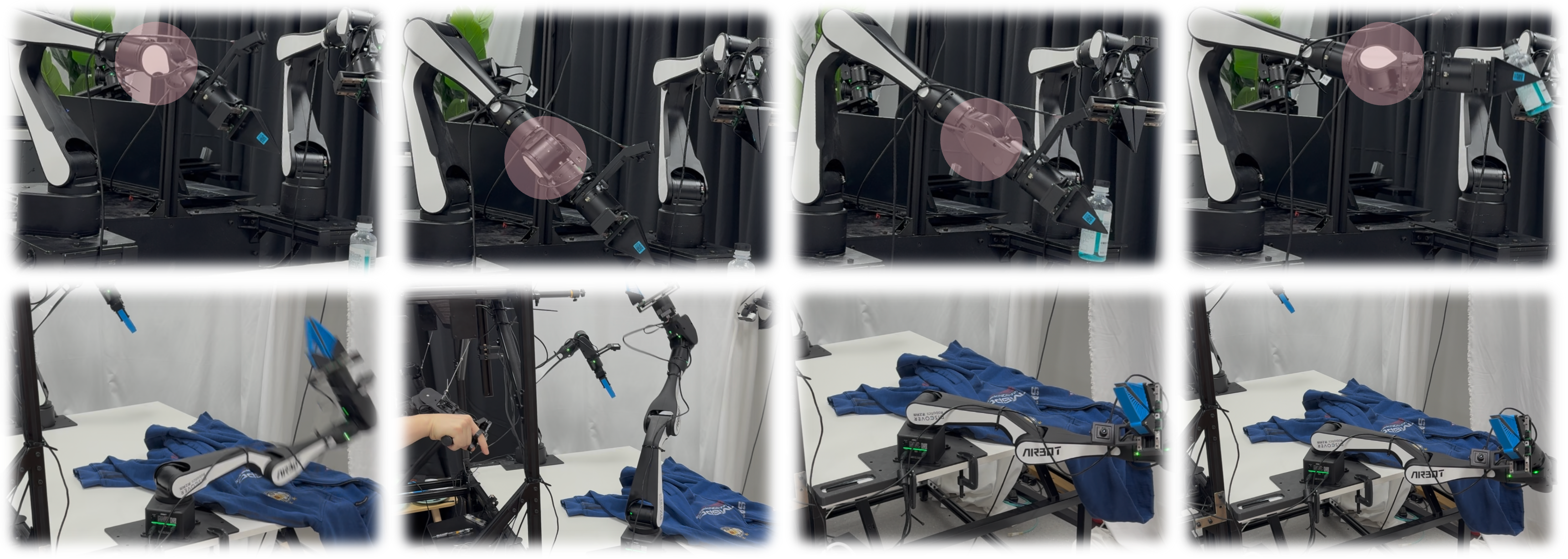}
    \caption{\textbf{Qualitative Comparison during Real-Robot OOD Deployment.} \textbf{(Top)} Traditional numerical solvers (e.g., cuRobo) fall into singularity traps near singular configurations, executing erratic wrist flips (red circles) that risk hardware collision. \textbf{(Bottom)} \textbf{MimicIK (Ours)} leverages conditional flow matching to inject human-like "self-rescuing" priors, effortlessly navigating the same trajectory with strict temporal smoothness and no explosive joint rotations.}
    \label{fig:real_robot_deployment}
\end{figure}
\vspace{-5mm}

\section{Discussion and Limitations}
\vspace{-2mm}
\label{sec:discussion & Limitations}

	We presented MimicIK, a real-time generative inverse kinematics solver that bridges embodiment-agnostic policies and physical hardware. By utilizing conditional delta-joint flow-matching and a two-step Minimal Iterative Policy (MIP), MimicIK filters teleoperation noise without the latency of standard diffusion. Additionally, our differentiable FK consistency loss structurally anchors predictions, preventing divergence and enabling autonomous singularity recovery.
    
    Despite these results, several limitations remain. First, MimicIK is currently tied to specific training hardware (the 6-DOF AIRBOT). Extending this to a cross-embodiment architecture—perhaps via dynamic URDF features or kinematic graphs—is a critical next step. Second, while inference is fast (6.74 ms at 20 Hz), it still relies on GPUs. Distilling the model for edge deployments with strict thermal budgets remains an open challenge. Finally, performance heavily relies on the teleoperation data quality; extreme distribution shifts could introduce motion artifacts. Future work will explore test-time guidance for real-time obstacle avoidance and self-collision penalties.




\bibliography{example}  

\newpage
\section{APPENDIX}
\label{appendix}

\subsection{A}\textbf{Real-Robot Deployment and Singularity Escapes}
\label{app:real_robot}

This appendix provides extended quantitative details regarding the physical deployment environment discussed in Section 4.4, acting as a textual companion to the supplementary video.

\subsection{Hardware Safety Limits and MLP Divergence}

\begin{wrapfigure}{r}{0.260\linewidth}
    \centering
    \includegraphics[width=\linewidth]{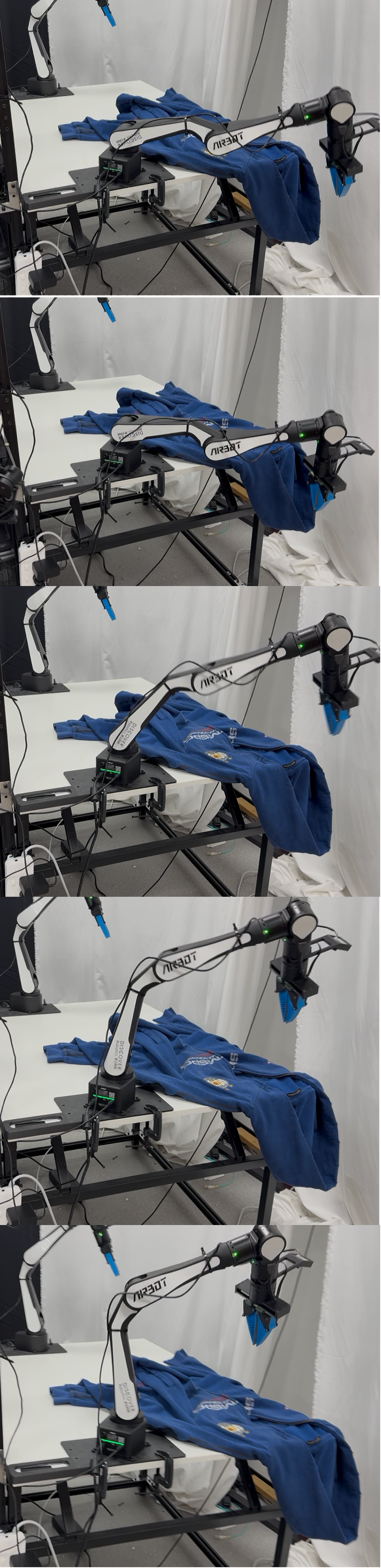}
    \caption{\raggedright \textbf{MIP Singularity Recovery.} MimicIK gracefully navigates out of kinematic dead-ends with strict temporal smoothness, avoiding erratic wrist flips.}
    \label{fig:mip_recovery}
\end{wrapfigure}

Physical deployment imposes stringent safety and continuity constraints that are absent in offline Cartesian evaluation. The 6-DOF AIRBOT platform used in our experiments operates at a 20 Hz closed-loop control frequency (a 50 ms cycle budget). At the hardware level, the URDF model does not enforce soft velocity limits; instead, low-level motor controllers enforce strict over-velocity protective limits. Specifically, instantaneous angular displacements exceeding $0.1$ rad/frame ($\approx 115^\circ/\text{s}$) trigger mild velocity warnings, while displacements exceeding $0.5$ rad/frame ($\approx 573^\circ/\text{s}$) immediately trigger a protective hardware E-stop (Emergency Stop).

As noted in the main text, the Pure MLP baseline completely failed this physical deployment test. In our out-of-distribution (OOD) robustness suite (comprising three long-horizon trajectories near singularities totaling $\sim$24,895 steps), all MLP seeds catastrophically diverged. Under closed-loop autoregressive deployment (where the previous step's prediction serves as the current state), the pure MLP outputs unconstrained predictions that rapidly explode into \texttt{NaN} values within approximately $310$ to $333$ frames. This triggers an irreversible "NaN-propagation" loop, locking the system. Notably, appending the FK loss to the MLP baseline during training provided no resilience against this OOD divergence, highlighting that geometric soft-constraints are insufficient without a robust generative prior.

\subsubsection{Anatomy of a Singularity Trap: The "180-Degree Wrist Flip"}
To understand why traditional numerical solvers (e.g., cuRobo, Pinocchio) are hazardous for continuous hardware deployment despite their near-perfect offline metrics, we analyzed their stepwise joint displacements.

When evaluated with warm-starting (using the previous frame as the initialization), the Pinocchio solver achieves a seemingly flawless offline mean error of $0.005$ mm. However, near singular configurations—where the Jacobian matrix becomes rank-deficient—the unconstrained pseudo-inverse solver mathematically forces the distal joints to instantaneously flip branches to maintain a straight Cartesian line. 

Our logs reveal that during these branch switches, Pinocchio commands a catastrophic maximum joint displacement ($\Delta q_{\text{max}}$) of \textbf{$243.6$ rad/frame}. At 20 Hz, this equates to an impossible angular velocity of $4,872$ rad/s (approx. $279,100^\circ/\text{s}$), physically demanding the robot to rotate 39 full revolutions within 50 ms. Even more severely, when deployed without warm-starting, the maximum displacement spikes to $814,399$ rad/frame, with $67.69\%$ of the trajectory exceeding the $0.1$ rad/frame safety threshold. These "wrist flips" mathematically minimize Cartesian error but guarantee violent hardware collisions on a physical manipulator.

\subsubsection{Extended Qualitative Results and Video Guide}
By formulating inverse kinematics as a conditional flow-matching process trained on human demonstrations, MimicIK learns a strict "self-rescuing" prior. When human operators encounter near-singular traps, they naturally relax strict Cartesian tracking to preserve joint-space smoothness.

As demonstrated in the supplementary video (and supported by our error ablation logs on \texttt{episode\_000009}), MimicIK inherits this behavior. MimicIK restricts the maximum joint displacement to a safe $0.12$ rad/frame (preventing E-stops) and maintains a bounded trajectory spike rate of only $1.67\%$ at the $0.1$ rad threshold. Instead of explosive branch switching, MimicIK smoothly transitions through singularities by slightly relaxing task-space tracking precision, heavily prioritizing the temporal smoothness of the joints before swiftly recovering the target pose.

\begin{figure}[htbp]
    \centering
    \includegraphics[width=\columnwidth]{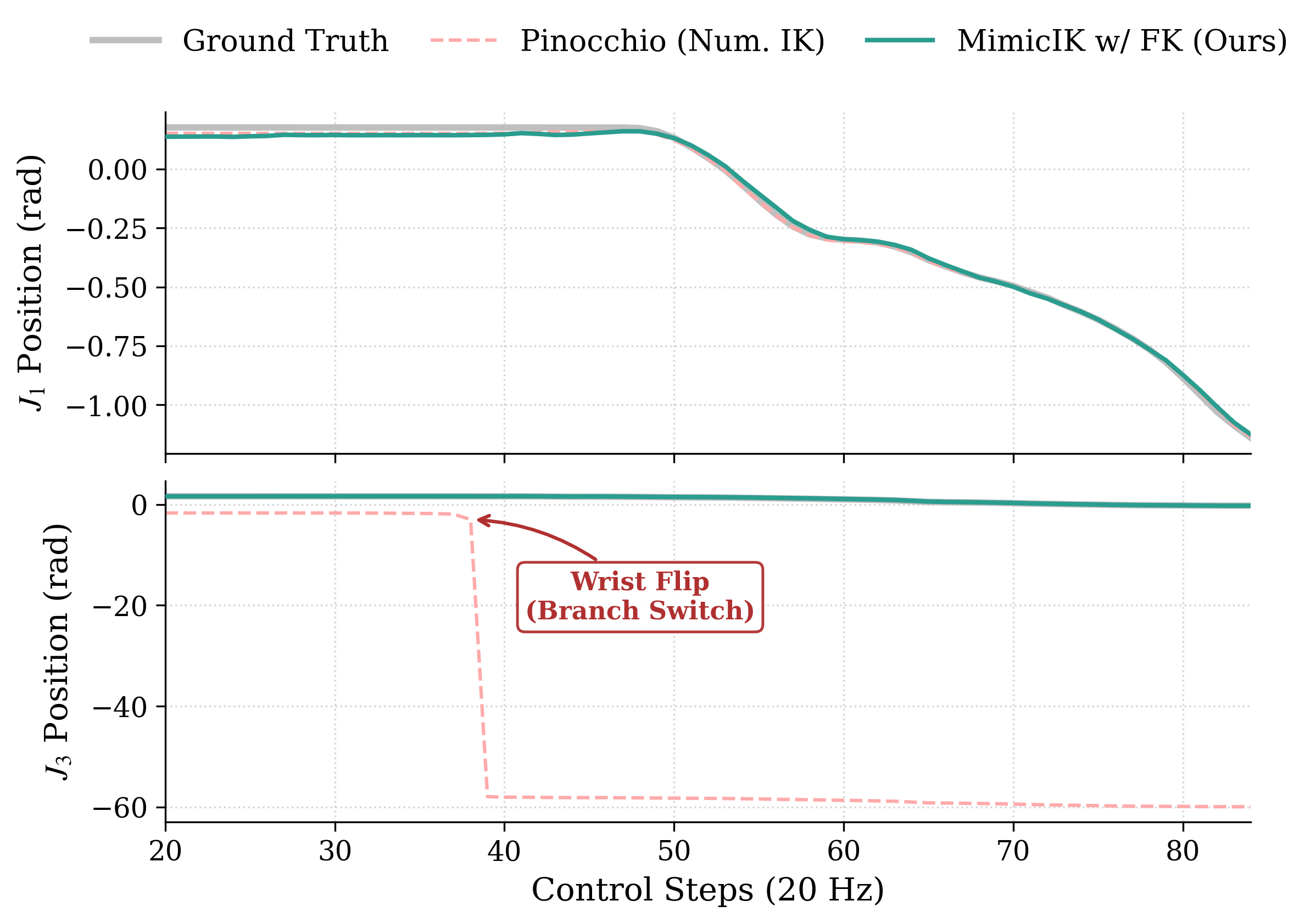}
    \caption{\textbf{Joint-Space Error and Smoothness Comparison.} Trajectory analysis comparing our generative MimicIK against a traditional numerical solver (Pinocchio). Around control step 38, the unconstrained numerical solver triggers a catastrophic branch switch (a "wrist flip" dropping by nearly 60 radians instantaneously) to minimize local Cartesian error, which would inevitably trigger hardware E-stops. In contrast, MimicIK learns to prioritize temporal continuity, perfectly tracking the ground truth without hazardous joint spikes.}
    \label{fig:trad_vs_mimicik}
\end{figure}

\subsection{B}\textbf{section{Ablation on FK Consistency Weight ($\lambda$)}}
\label{app:fk_ablation}

To determine the optimal regularization strength for the differentiable forward kinematics penalty, we conducted a systematic ablation study on the FK consistency weight, $\lambda \in \{0.05, 0.1, 0.2, 0.5\}$. The results, evaluated on the held-out test set, are summarized in Table~\ref{tab:fk_ablation}.

\begin{table}[htbp] 
    \centering
    \caption{\textbf{Quantitative Ablation on the FK Consistency Loss Weight ($\lambda$).} Evaluated on the held-out test set. The model achieves the optimal balance of spatial accuracy and motion smoothness at $\lambda = 0.1$.}
    \label{tab:fk_ablation}
    \resizebox{\columnwidth}{!}{%
    \begin{tabular}{lccccc}
        \toprule
        $\lambda$ Weight & Mean Err. (mm) $\downarrow$ & P95 (mm) $\downarrow$ & SR@5mm (\%) $\uparrow$ & SR@10mm (\%) $\uparrow$ & Spike Rate (\%) $\downarrow$ \\
        \midrule
        0.05 & 5.00 & 12.26 & 56.20 & 91.42 & 8.58 \\
        \rowcolor{gray!15} \textbf{0.10 (Ours)} & \textbf{4.65} & 11.16 & \textbf{76.40} & 92.01 & 7.99 \\
        0.20 & 4.70 & \textbf{9.46} & 64.20 & \textbf{92.34} & \textbf{7.66} \\
        0.50 & 6.05 & 13.68 & 35.30 & 90.88 & 9.12 \\
        \bottomrule
    \end{tabular}%
    }
\end{table}

The ablation results clearly demonstrate a trade-off between geometric constraint and generative freedom:
\begin{itemize}
    \item \textbf{Optimal Balance ($\lambda=0.1$):} This setting acts as the "sweet spot", achieving the lowest mean position error (4.65 mm) and the highest stringent success rate (SR@5mm of 76.40\%). It provides sufficient structural guidance without stifling the learned motion priors.
    \item \textbf{Under-regularization ($\lambda=0.05$):} When the geometric constraint is too weak, the error rebounds to 5.00 mm, and the fine-grained accuracy (SR@5mm) drops significantly to 56.20\%, indicating insufficient physical anchoring.
    \item \textbf{Moderate Over-regularization ($\lambda=0.2$):} While a stronger penalty improves the worst-case error (P95 improves to 9.46 mm) and trajectory smoothness (Spike rate drops to 7.66\%), the rigid constraint begins to restrict the generative model's fine-grained precision, causing SR@5mm to drop to 64.20\%.
    \item \textbf{Severe Over-regularization ($\lambda=0.5$):} At this level, the FK constraint aggressively dominates the training objective. Consequently, SR@5mm collapses to 35.30\%, and the mean error rises to 6.05 mm, as the model struggles to balance the conflicting gradients from the joint-space flow matching and the task-space FK loss.
\end{itemize}

\subsection{C}\textbf{Network Architecture and Training Details}
\label{app:hyperparams}

This appendix provides complete implementation details to support the reproducibility of our results.

\subsubsection{SuDeepDiT Architecture}
The conditional flow generator in MimicIK is based on SuDeepDiT, a symmetric encoder-decoder Transformer. To ensure real-time efficiency, the architecture encodes the observation history $[q_\text{curr}, \text{EE}_\text{curr}, \text{EE}_\text{tgt}]$ and time scalars $(s, t)$ using standard MLPs and sinusoidal embeddings. These condition a core network of 8 Transformer encoder blocks and 8 DiT-style decoder blocks (which utilize Adaptive Layer Normalization with zero-initialization) to predict the continuous velocity field.

\begin{table}[htbp]
    \centering
    \caption{\textbf{Input and Output Dimensions of the Flow Generator}}
    \begin{tabular}{llc}
        \toprule
        \textbf{Symbol} & \textbf{Description} & \textbf{Dimension} \\
        \midrule
        $\mathbf{x}_t$ & Noisy action (joint delta) at flow time $t$ & $(B, T_a, 6)$ \\
        $s, t$ & Flow time scalars (source and target) & $(B,)$ each \\
        $\mathbf{o}$ & Observation history $[q_\text{curr}, \text{EE}_\text{curr}, \text{EE}_\text{tgt}]$ & $(B, T_o, 20)$ \\
        $\hat{\mathbf{v}}$ & Predicted velocity field & $(B, T_a, 6)$ \\
        \bottomrule
    \end{tabular}
    
    \vspace{3mm}
    
    \caption{\textbf{Architecture Summary of SuDeepDiT}}
    \begin{tabular}{ll}
        \toprule
        \textbf{Hyperparameter} & \textbf{Value} \\
        \midrule
        Hidden dimension $d$ & 256 \\
        Encoder / Decoder layers $N$ & 8 / 8 \\
        Attention heads & 4 \\
        FFN dimension & 2048 ($8\times d$) \\
        Time embedding dim & 128 (per scalar) \\
        Total parameters & $\sim$16.2 M (flow map) + 0.2 M (encoder) \\
        \bottomrule
    \end{tabular}
\end{table}

\subsubsection{Training Hyperparameters}
All MimicIK (MIP + FK) models are trained using the AdamW optimizer with a learning rate of $1 \times 10^{-4}$ and a cosine annealing schedule (no warmup).

\begin{table}[htbp]
    \centering
    \caption{\textbf{Training and Optimization Details}}
    \begin{tabular}{ll}
        \toprule
        \textbf{Setting} & \textbf{Value} \\
        \midrule
        Optimizer & AdamW (Weight decay $= 1 \times 10^{-5}$) \\
        Gradient clipping & Max norm $= 10.0$ \\
        Batch size & 512 \\
        Total training steps & 300,000 \\
        EMA decay rate & 0.995 (Updated every step) \\
        Hardware & 1$\times$ NVIDIA RTX 4080 SUPER (16 GB) \\
        Training time per run & $\sim$13 hours ($\sim$6.4 steps/sec) \\
        \bottomrule
    \end{tabular}
\end{table}

\textbf{Data Normalization:} For observations ($q_\text{curr}$, EE poses), joint angles and EE positions use linear min-max normalization to $[-1, 1]$. Quaternion components pass through an identity normalizer after a half-sphere alignment step ($q_w \ge 0$) to eliminate double-cover ambiguity. For actions, the joint delta $\Delta q$ also uses min-max normalization to $[-1, 1]$. Normalizer statistics are fit exclusively on the training split.

\subsubsection{Differentiable FK Consistency Loss Implementation}
The FK consistency loss is computed in a separate forward-backward pass from the Minimal Iterative Policy (MIP) flow-matching loss to strictly avoid gradient interference. Specifically, after predicting the terminal action $\hat{\mathbf{x}}_1$, we denormalize it to obtain the joint delta $\widehat{\Delta q}$ and compute the commanded joint configuration $\hat{q} = q_\text{curr} + \widehat{\Delta q}$. The EE position and rotation are then computed via a differentiable forward kinematics engine (\texttt{pytorch\_kinematics}).

The two backward passes share the same optimizer but use separate \texttt{zero\_grad()} calls. This engineering choice guarantees that gradients from the task-space geometric penalty do not inaccurately accumulate or interfere with the joint-space flow matching objective.
\end{document}